\documentclass[conference]{IEEEtran}
\pdfoutput=1
\IEEEoverridecommandlockouts
\usepackage{cite}
\usepackage{amsmath,amssymb,amsfonts}
\usepackage{algorithmic}
\usepackage[pdftex]{graphicx}
\usepackage{epstopdf}
\usepackage{textcomp}
\usepackage{xcolor}
\usepackage{bbding}
\usepackage{subfigure}
\usepackage{threeparttable}
\definecolor{colorfirst}{rgb}{.866,.945, 0.831} 
\definecolor{colorsecond}{rgb}{1, 0.98, 0.83} 
\definecolor{colorthird}{rgb}{1, 1, 1} 
\definecolor{bronze}{rgb}{1,1,0.6}
\definecolor{silve}{rgb}{0.969,0.796,0.600}
\definecolor{gold}{rgb}{0.941,0.592,0.600} 

\newcommand{\gold}[1]{\colorbox{gold}{{#1}}}
\newcommand{\silve}[1]{\colorbox{silve}{{#1}}}

\usepackage{bm}
\def\BibTeX{{\rm B\kern-.05em{\sc i\kern-.025em b}\kern-.08em
    T\kern-.1667em\lower.7ex\hbox{E}\kern-.125emX}}

\usepackage[bookmarks=false]{hyperref}
\usepackage{cleveref}
\usepackage{siunitx} 
\usepackage{tabularx}

\usepackage{multirow}

\usepackage{enumitem}
\usepackage{bm}
\usepackage{color}
\usepackage{makecell}
\usepackage{setspace}

\renewcommand{\footnoterule}{%
  \kern -3pt
  \hrule width 0.2\textwidth height 0.5pt
  \kern 2pt
} 
\begin{document}

\title{Fast and Physically-based Neural Explicit Surface for Relightable Human Avatars
}


\author{
	\textit{Jiacheng Wu$^{1*}$, Ruiqi Zhang$^{1*}$, Jie Chen$^{1\dag}$, Hui Zhang$^{2}$}\\
	$^1$Department of Computer Science, Hong Kong Baptist University, Hong Kong SAR, China \\
    $^2$Department of Computer Science, Beijing Normal-Hong Kong Baptist University, 
    Guangdong, China \\ 
    \tt\small Email: \{csjcwu, csrqzhang, chenjie\}@comp.hkbu.edu.hk, amyzhang@uic.edu.cn \\
    \thanks{$^*$Equal contribution. $^\dag$Corresponding author.}
}

\maketitle
\renewcommand{\thefootnote}{}

\begin{abstract}
Efficiently modeling relightable human avatars from sparse-view videos is crucial for AR/VR applications. Current methods use neural implicit representations to capture dynamic geometry and reflectance, which incur high costs due to the need for dense sampling in volume rendering. To overcome these challenges, we introduce Physically-based Neural Explicit Surface (PhyNES), which employs compact neural material maps based on the Neural Explicit Surface (NES) representation. PhyNES organizes human models in a compact 2D space, enhancing material disentanglement efficiency. By connecting Signed Distance Fields to explicit surfaces, PhyNES enables efficient geometry inference around a parameterized human shape model. This approach models dynamic geometry, texture, and material maps as 2D neural representations, enabling efficient rasterization. PhyNES effectively captures physical surface attributes under varying illumination, enabling real-time physically-based rendering. Experiments show that PhyNES achieves relighting quality comparable to SOTA methods while significantly improving rendering speed, memory efficiency, and reconstruction quality.
\end{abstract}


\section{Introduction}
Reconstructing dynamic human avatars with realistic physical attributes is crucial for applications in computer games and 3D movies. Accurately depicting human models in 3D environments requires detailed modeling and effective simulation of physical attributes, posing significant challenges. 
Traditional methods use expensive capture devices, such as controllable illumination matrices~\cite{guo2019relightables, habermann2019livecap} and dense camera rigs~\cite{gortler1996lumigraph}, whose generalization is limited by necessary specialized equipment and manual labor. Recent advancement in this area involves using neural fields, like the Neural Radiant Field (NeRF) model~\cite{mildenhall2020nerf}, which employs neural networks to capture density and color information from sparse-view images. Following NeRF's success, implicit representations~\cite{wang2021neus} have become popular for 3D avatar reconstruction, enabling cost-effective modeling of dynamic avatars from image observations. 
Some advancements~\cite{xu2022surface, zhang2022ndf} focus on learning dynamic implicit representations conditioned on human poses, and some~\cite{peng2024animatable} further incorporate Signed Distance Fields to leverage surface constraints and improve mesh reconstruction. Despite these improvements, the rendering efficiency of implicit representations remains an issue, as volumetric rendering requires numerous network queries, leading to high computational demands and restricted applications. To address this, we propose an approach that learns neural maps for dynamic human models and employs a rasterization-based neural rendering mechanism for real-time performance.
Alongside implicit methods, explicit representations like meshes and 3D Gaussian splatting have been used for dynamic human avatar modeling~\cite{qian20233dgs, zhang2024mesh}, but they often face memory issues at higher resolutions. Parametric models like SMPL~\cite{loper2015smpl} excel in body modeling but struggle with clothed avatars, which typically need costly 3D scans. PhyNES addresses this by modeling pose-dependent deformations as surface changes over a parametric model and integrating neural maps, enabling efficient training from sparse-view videos. This approach optimizes storage by incorporating surface offsets and texture maps into the parametric model, significantly reducing storage demands.

Research in the physically-based rendering of human avatars has significantly advanced, particularly in creating realistic representations from sparse video input under given illumination~\cite{peng2024animatable, liu2021neural}.
However, these methods often struggle with lighting variations, as fixed elements like shadows hinder adaptability to dynamic conditions, making it essential to disentangle physical characteristics in the computer graphics pipeline to enhance versatility.
Prior techniques, such as photometric stereo~\cite{habermann2019livecap} and controllable illumination arrays~\cite{ guo2019relightables, yang2023towards}, can restore high-quality human materials and yield excellent relighting results but typically require professional setups, limiting their broader applicability. Consequently, learning-based techniques are essential for extracting material properties from photos and videos. Some learning-based neural inverse rendering approaches~\cite{alldieck2022photorealistic, yeh2022learning} aim to predict these properties from a single image. However, they often face challenges in surface reconstruction due to insufficient geometric constraints and their primary design for static scenes, which makes them less effective in dynamic environments.
Therefore, neural rendering techniques have been proposed to help produce relightable models grounded in geometric assumptions. Approaches like Relighting4D~\cite{chen2022relighting4d} and Relightable Avatar (RA)~\cite{xu2024relightable} have attempted to rebuild human materials from sparse views. While these techniques advance lighting modeling, they rely on neural implicit representations, which are computationally intensive and incur substantial training costs. To tackle these challenges, our PhyNES employs material disentanglement through an explicit mesh surface, utilizing pose-dependent normal texture mapping to enhance surface detail and allow precise surface information querying. This mesh-based approach optimizes visibility assessment across various poses while leveraging pose-conditioned neural fields to enhance the overall fidelity of avatars.

Our proposed PhyNES focuses on reconstructing dynamic human models with physical characteristics from sparse-view videos using pose-conditioned neural fields~\cite{zhang2022ndf, xu2022surface, zhang2023explicifying}. 
We model a human as several neural material maps, implemented by 2D hash-encoding MLPs, as pose-specific texture color maps and pose-dependent physical parameter maps. 
We start by utilizing Neural Explicit Surface (NES)~\cite{zhang2023explicifying} and enhancing training efficiency with hash encoding~\cite{muller2022instant}. 
Defining as an explicit surface allows rasterization to be incorporated to project mesh to the image plane. Thus, we only query MLP once for each pixel, efficiently improving inference speed compared to volumetric rendering methods.
We use neural maps employed by a 2D UV-coordinate-based MLP to disentangle physical attributes. 
Visibility and lighting manipulation 
can be subsequently performed efficiently on a defined mesh surface. 
Storing an animatable avatar efficiently as a combination of SMPL parameters and 2D surface maps containing texture, geometry, and material information enhances memory and inference efficiency for real-time applications with PhyNES. This paper makes several key contributions:

\hspace{-0.4cm}$\bullet$ We introduce PhyNES to enhance the sparse-view dynamic avatar modeling framework based on NES for physically-based rendering and relighting applications. Our fully differentiable framework effectively disentangles dynamic avatar geometry, surface texture, and material information from environmental illumination using sparse-view video input. 
\\$\bullet$ We propose a rasterization-based neural renderer that efficiently rasterizes the $uv$ coordinates for all on-screen pixels for parallel querying of geometry, texture, and material networks, significantly improving rendering and relighting efficiency.
\\$\bullet$ We perform comprehensive experiments to confirm PhyNES' superior efficiency in dynamic human modeling and relighting compared to existing frameworks.

\vspace{-0.1cm}
\section{The Proposed Method}
\label{sec: method}
\begin{figure*}[ht!]

\setlength{\abovecaptionskip}{-0.cm}
  \centering
  \resizebox{1\linewidth}{!}
  {
  \includegraphics[width=1.\linewidth]{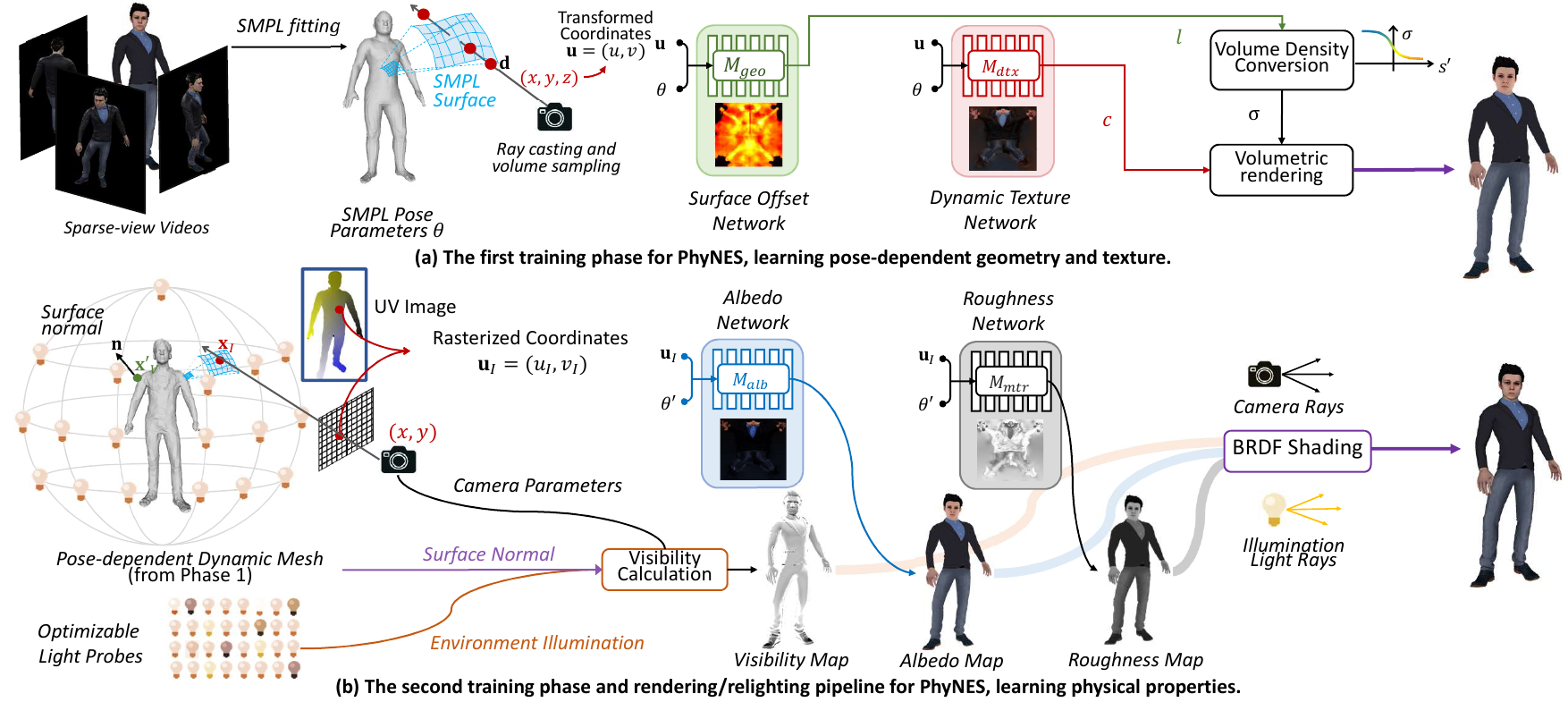}
   }
   \vspace{-0.5cm}
  \caption{
  The overall rendering and relighting pipeline of PhyNES includes two learning phases, as shown in (a) and (b). (a) In the initial stage, the fitted SMPL model will be used as a reference plane to transform volume sampling points' world coordinates to a thin layer of transformed $uv$ coordinate space. $uv$ and the SMPL pose parameters $\theta$ are used as input to a hash surface offset network and a dynamic texture network. These estimators predict the offset $l$ and color $c$ at the corresponding UV coordinates, and the signed distance is computed using a conversion module. A pose-dependent dynamic mesh (both in surface offset and textures) will be generated. (b) An optimizable light probe array will be configured around the stage in the second phase. The surface normals can be conveniently computed with the rasterization-based neural renderer for each learnable probe, incidence, and observation direction. An albedo and roughness network will predict respective surface material attributes for each rasterized coordinate $uv$ to facilitate relighting applications. Our model produces the final physically-based rendering output by connecting these attributes to a differentiable BRDF~\cite{cook1982reflectance} function.}
  \vspace{-0.6cm}
  \label{fig:overview}
\end{figure*}

We present PhyNES, an efficient framework that captures the dynamic appearances of human avatars from sparse-view videos
, and enables physically-based relighting applications. 
As shown in Fig.~\ref{fig:overview}, PhyNES comprises two training stages: the first focuses on learning the pose-dependent geometry and texture, while the second emphasizes the physically-based properties necessary for relighting applications. 
Sec.~\ref{sec: representation} provides an overview of the Neural Explicit Surface (NES) representation and our improvements. Then, Sec.~\ref{sec: rendering} describes our rasterization-based neural material renderer, enabling efficient querying of neural maps employed by material networks.
Finally, Sec.~\ref{sec: pbr} details the Physically-based Rendering process.

\setlength{\textfloatsep}{0pt}

\subsection{Preliminary: Neural Explicit Surface}
\label{sec: representation}

We build our method on Neural Explicit Surface (NES)~\cite{zhang2023explicifying} for its efficiency in modeling human surfaces, which is crucial for PBR. 
Specifically, NES models human surfaces using fitted SMPL~\cite{loper2015smpl} as surface offset and texture, implemented by 2D UV-coordinate-based MLP. The kernel of NES is a differentiable Signed Distance Conversion Module, which converts the signed distance from a sample point to an SMPL surface as its signed distance from the real surface. In Fig.~\ref{fig:overview}(a), a sample point $(x, y, z)$ is projected to the SMPL surface to derive UV coordinates $(u,v)$ of the closest point, which is input to Surface Offset Network $M_{geo}$ and Dynamic Texture Network $M_{dtx}$. These networks predict surface offset value $\bm{l}$ and texture color $c$, where $\bm{l}$ indicates the distance from the SMPL surface to the actual surface along the SMPL surface normal. The signed distance $\bm{s}$ of a sample point can be approximated with $\bm{s\!=\!h\!-\!l}$, where $\bm{h}$ is the distance from the sample point to the SMPL surface. This conversion module enables NES to be optimized with volume rendering~\cite{kajiya1984ray}, enhancing effective 3D reconstruction from multi‐view images. 
Essentially, NES transforms costly implicit representations into two efficient pose-dependent maps using neural networks. The queried texture color $c$ and offset value $\bm{l}$ can be mapped back to the 3D space, resulting in color and density values for volumetric rendering. This end‐to‐end trainable setup disentangles surface geometry from texture using sparse‐view observations, improving compatibility with graphics pipelines and opening avenues for modeling other material dynamics, paving the way for the extension of NES to PhyNES for physically-based rendering.

\textbf{Improvements.} We enhance NES by replacing the original regular 2D UV-coordinate-based MLP with hash-encoding MLP~\cite{muller2022instant} to boost efficiency and performance. Additionally, we adopt mesh loss $l_{mesh}$~\cite{ravi2020pytorch3d}, including mesh edge loss, mesh normal smoothness loss, and mesh Laplace loss, to regularize and enhance reconstruction mesh.

\subsection{Rasterization-based Neural Material Renderer}
\label{sec: rendering}
The implicit representation of illumination disentanglement is slow.
To address this, we harness NES's efficiency to explicitly model PBR and separate surface material information using an efficient mesh representation.
Mesh representation allows us to quickly find the intersection points between rays and the human body surface to calculate the relationship between rays and normals to obtain visibility. It also enables rapid determination of ray occlusion to capture occlusion relationships.
Following how NES has disentangled surface geometry and dynamic texture as separate 2D neural maps with two neural networks ($M_{geo}$ and $M_{dtx}$), PhyNES aims to further disentangle the neural texture map into a 2D albedo map and a roughness map with two additional networks, i.e. the albedo network $M_{alb}$ and roughness network $M_{rgh}$, from sparse view videos in a differentiable end-to-end manner. 
As shown in Fig.~\ref{fig:overview}(b), both $M_{alb}$ and $M_{rgh}$ take the $(u,v)$ coordinates and the current SMPL pose parameter $\theta$ to output corresponding albedo and roughness values. Essentially, PhyNES seeks to decompose the dynamic texture network $M_{dtx} $'s outputs with these two material networks $M_{alb}$ and $M_{rgh}$. These material variations interact with the camera and lighting configuration to produce the final shading, and this shading process is governed by the Bidirectional Reflectance Distribution Function (BRDF)~\cite{walter2007microfacet} introduced in Sec.~\ref{sec: pbr}. 

Querying efficiency is a critical consideration due to the need for additional networks at each sampling point. However, the explicit nature of NES allows us to utilize a rasterization‐based neural renderer to address this issue effectively. To determine the material attributes of each pixel on the screen, we employ the rasterization from Pytorch3D. Each pixel is assigned a texel coordinate $(u, v)$ based on the ﬁrst intersection of a tracing ray with the deformed mesh, resulting in a UV image as shown in Fig.~\ref{fig:overview}(b) that records the texel coordinate $\textbf{u}=(u, v)$ for each image‐space coordinate $(x, y)$. This highly efficient graphics pipeline operation enables parallel calculations for all on‐screen pixel texel coordinates, thus facilitating efficient parallel queries to the material networks. 
Except for albedo and roughness, this renderer efficiently retrieves normal information from the deformed mesh, which is essential for assessing visibility from the current camera angle. The rasterization‐based neural material renderer thus enhances rendering efficiency and reduces computational demands through its ability to process all pixels in parallel.

\vspace{-0.1cm}{
\subsection{Physically-based Avatar Relighting}
\label{sec: pbr}
}
We explicitly model the environment illumination with a matrix of learnable probes $\bm{L_s}$ to enable physically-based avatar relighting. As shown in Fig.~\ref{fig:overview}(b), these probes are distributed uniformly over a sphere that encircles the pose-dependent dynamic mesh to simulate the illumination on site. Light rays emitted from each light probe interact with the mesh surface, resulting in the computation of the final outgoing radiance $\bm{L_o}$ corresponding to the observation angle $\bm{\omega_o}$~\cite{kajiya1986rendering}:

\vspace{-0.1cm}
{\scriptsize
\begin{align}
\setlength{\abovedisplayskip}{0pt}
\setlength{\belowdisplayskip}{0pt}
\hspace{-0.225cm}\bm{L_o}(x_s,\!\bm{\omega_o})\!=\!\int_\Omega\!\bm{L_s}(\bm{\omega_i})\!\cdot\!R_s(x_s,\!\bm{\omega_i},\!\bm{\omega_o},\!\bm{n_s})\!\bm{\cdot}\!
V_s(x_s,\!\bm{\omega_i})\!\cdot\!(\bm{\omega_i}\!\cdot\!\bm{n_s})d\omega_i.
\label{eq:brdf}
\end{align}
}

Here, $x_s$ denotes the intersection point between the incoming light ray $\bm{L_s(\omega_i)}$ and the surface, and $\bm{n_s}$ denotes the surface normal at $x_s$, which can be queried via the neural rendering method of rasterization as introduced previously. $V_s(x_s,\!\bm{\omega_i})$ denotes the visibility map representing the occlusion relationships between $\bm{L_s}$ and $x_s$, as well as the angle between $\bm{n_s}$ and $\bm{\omega_i}$ at $x_s$. Specifically, we decide self-occlusion by emanating light from the mesh surface along the normal direction to judge whether it is occluded by self and calculate the angle by multiplying surface normals and rays. $R_s(x_s,\!\bm{\omega_i},\!\bm{\omega_o},\bm{n_s})$ corresponds to the BRDF function, and we adopted the Microfacet model~\cite{walter2007microfacet}. It utilizes the Cook-Torrance kernel~\cite{cook1982reflectance} $f_r$, which partitions the object's surface reflection into diffuse and specular components according to: 

{\footnotesize
\begin{align}
f_r = k_d f_\text{Lambert} + k_s f_\text{Cook-torrance},
\label{eq:nonlinear}
\end{align}}\label{eq:cook}
where $k_d$ and $k_s$ denote the ratio of energy in incident light that are partitioned to the diffusive and specular components, respectively. Here, we denote these two items as 1. The diffusive $f_\text{Lambert}$ is computed from queried $\alpha_s$ from $M_{alb}$, while the specular part $f_\text{Cook-Torrance}$ is calculated from queried $\gamma_s$ from $M_{rgh}$ in~\eqref{eq:query}. The queried process is shown as~\eqref{eq:query}:

{\footnotesize
\begin{align}
M_{alb}((u, v), \theta) = \alpha_s,\notag
\\M_{rgh}((u, v), \theta) = \gamma_s.
\label{eq:query}
\end{align}
}

Leveraging the differentiable rendering equation in~\eqref{eq:brdf}, PhyNES efficiently disentangles the albedo and roughness maps by learning from sparse view projections over the rendering equation. This process, powered by the highly efficient rasterization-based material renderer, enables us to relight the avatar under various lighting conditions by re-configuring $L_s$.

\textbf{Remark.} Compared with other methods focused on relighting human avatars~\cite{xu2024relightable, chen2022relighting4d}, our approach stands out for its use of texel coordinates to query normal and learn material attributes in 2D space, leading to notable gains in memory efficiency. This strategic approach reduces memory redundancy and computational burden, culminating in a more streamlined and resource-efficient neural rendering workflow.

\section{Experiments}
\label{sec: experiment}
We collect three datasets for training and evaluation:
(1) \textit{ZJU-MoCap}~\cite{peng2021neural} is a real-world dataset (23 views) dataset on 6 humans. We select five individuals and four random views for training and evaluating novel pose and view rendering. 
(2) \textit{SyntheticHuman}~\cite{peng2024animatable} contains 7 dynamic 3D human models (10 views) with ground truth mesh. We select seven models and four views to train and assess mesh reconstruction quality.
(3) \textit{SyntheticHuman++}~\cite{xu2024relightable} includes 6 sequences (20 views) of dynamic 3D human models with ground truth relighting information. We choose six persons and four views for training and comparison with existing relighting methods.

To evaluate geometry reconstruction quality, we measure \textbf{Chamfer Distance (CD)} for point cloud similarity and \textbf{Point-to-Surface Distance (P2S)} for distances from points to the nearest reference surface. Besides, we use \textbf{PSNR and LPIPS} for image quality assessment.

\subsection{Seen- \& Novel- Pose Rendering Quality}
\begin{figure}[h]
\setlength{\abovecaptionskip}{-0.2cm}
\centering
\includegraphics[width=0.9\linewidth]{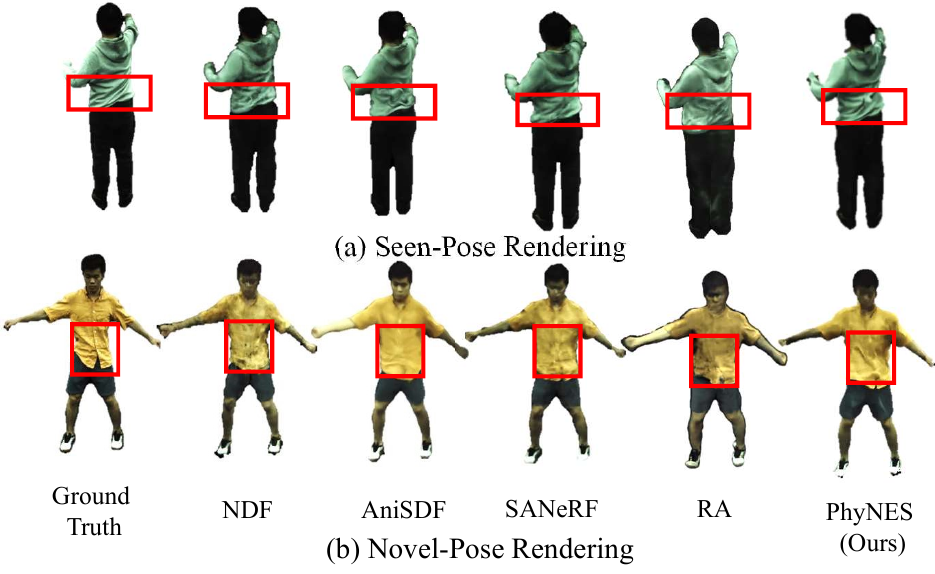}
\caption{Qualitative comparison in (a) seen-pose rendering and (b) novel-pose rendering tasks on the ZJU-MoCap dataset. As indicated in red squares, PhyNES can better model dynamic variations.}
\label{figure:novel_view}

\end{figure}

The qualitative comparison of rendering quality, as measured by PSNR and LPIPS, is presented in Table~\ref{table:novel_quan}. 
Note that we retrain RelightableAvatar (RA)~\cite{xu2024relightable} models, while results of NDF~\cite{zhang2022ndf}, AniSDF~\cite{peng2024animatable}, SANeRF~\cite{xu2022surface} and NES~\cite{zhang2023explicifying} are sourced from~\cite{zhang2023explicifying}. Experimental settings are identical among all comparisons.
In the seen-pose task, PhyNES shows slightly worse PSNR compared with others, but it significantly excels in the novel-pose task. This advantage is largely due to the efficient querying of texture and surface offsets in the compact 2D space, enhancing PhyNES's generalization capability.
We compare rendering performance with different methods for both seen and novel poses, as shown in Fig.~\ref{figure:novel_view}. 
This figure highlights that PhyNES captures pose-dependent cloth wrinkle variations more effectively than others, owing to its efficiency in modeling dynamic changes in a compact transformed space.

\begin{table}[t!]
\setlength{\abovecaptionskip}{-0.15cm}
\begin{center}
    \caption{Seen-pose and novel-pose rendering results on the ZJU-MoCap dataset. Best results are highlighted as \gold{1st} and \silve{2nd}.
    }
    \resizebox{0.85\linewidth}{!}{
    \begin{tabular}{c|c|c}
        & \multicolumn{1}{c|}{Seen-Poses} & \multicolumn{1}{c}{Novel-Poses}\\
    Avg. & PSNR (dB) $\uparrow$ / LPIPS $\times 10^3 \downarrow$ & PSNR (dB) $\uparrow$ / LPIPS $\times 10^3 \downarrow$\\
    \hline
    AniSDF   &   \silve{27.39} / 31.46            & \silve{24.90} / 36.80 \\
    NDF      &   26.94 / 28.68                    & 24.58 / 35.53    \\  
    SANeRF   &   \gold{27.56} / \silve{23.18}     & 24.72 / \silve{31.21} \\
    NES      &   27.20 / 23.68                    & 25.34 / 29.93 \\
    RA       &   27.31 / 27.43                    & 24.87 / 35.05 \\
    \hline
    Ours     &   27.26 / \gold{22.79}             & \gold{25.40} / \gold{29.58} \\
    \end{tabular}
    \label{table:novel_quan}
}

\end{center}
\end{table}

\subsection{Comparisons on Rendering Efficiency}
\label{mesh}

We evaluate rendering efficiency in terms of memory cost and rendering speed by testing different models, categorized as relightable or non‐relightable. Images are rendered at a resolution of $512 \times 512$ on a server equipped with 64 AMD EPYC 7302 16‐core processors and 2 A100 GPUs. The mean memory cost and rendering speed over 100 runs are provided in Table~\ref{table:efficiency}.
Both NES and PhyNES adopt a rasterization-based renderer, demonstrating signiﬁcantly higher speeds suitable for real‐time applications. The additional cost of PhyNES arises from its further material modeling and physically-based rendering calculations. In comparison, Relighting4D and RA, which also provide physical relighting for avatars, are significantly inferior to PhyNES in both efficiency metrics.

\begin{table}[t!]
\setlength{\abovecaptionskip}{0.cm}
\begin{center}
    \caption{Rendering evaluation results in terms of memory consumption (in Gigabytes (GB)) and rendering speed (in frames per second (FPS)). Best results are highlighted as \gold{1st} and \silve{2nd}.
    } 
    \resizebox{0.9\linewidth}{!}{
    \begin{tabular}{c|c|c|c}
        \textbf{Relightable}     & \textbf{Method}  & \textbf{Memory~(GB) $\downarrow$} &  \textbf{Speed (FPS)} $\uparrow$\\
    \hline
     & NDF~\cite{zhang2022ndf}                      & \silve{5.5}         & \silve{2.48}  \\
   No & AniSDF~\cite{peng2024animatable}             & 11.4        & 0.93  \\
    & NES~\cite{zhang2023explicifying}             & \gold{3.2}         & \gold{15.89} \\
    \hline
     & Relighting4D~\cite{chen2022relighting4d}     &  \silve{7.2}       &  \silve{0.33}       \\
  Yes  & RA~\cite{xu2024relightable}                  & 10.89        & 0.21   \\
    & PhyNES (Ours)                                      &\gold{3.5}         & \gold{12.27} \\
    \end{tabular}
    \label{table:efficiency}
}
\vspace{-0.2cm}
\end{center}
\end{table}

\subsection{Mesh Reconstruction Quality}
{
\vspace{-0.5cm}
\begin{figure}[h!]
\setlength{\abovecaptionskip}{0.cm}
\centering
\includegraphics[width=1.0\linewidth]{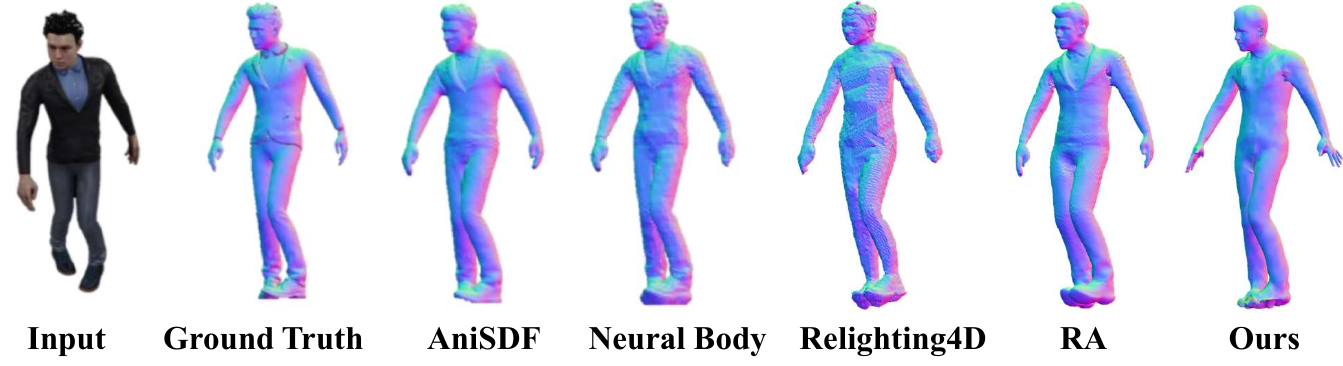}
\vspace{-0.5cm}
\caption{Comparative visualization of surface normals based on the reconstructed geometry by different methods on the SyntheticHuman dataset.}
\setlength{\textfloatsep}{0pt}
\label{fig:normal_qual}
\end{figure}
\vspace{-0.5cm}
}

{

\begin{table}[ht]
\begin{center}
\setlength{\abovecaptionskip}{-0.2cm}
    \caption{Results on Mesh Reconstruction on Synthetic Human Dataset. Best results are highlighted as \gold{1st} and \silve{2nd}.
    }
\setlength{\abovecaptionskip}{0.cm}
    \resizebox{1.0\linewidth}{!}{
    \begin{tabular}{c|c|c|c|c|c}
    P2S$\downarrow$ / CD$\downarrow$ & NB &  AniSDF & Relighting4D & RA & Ours \\
    
    
    \hline
    Avg.     & 1.10 / 1.16   & 0.59 / 0.73  & 1.37 / 1.33  &\gold{0.42} / \gold{0.52} & \silve{0.50} / \silve{0.62}\\
   
    \end{tabular}}
\vspace{-0.5cm}
    \label{table:mesh}
\end{center}
\end{table}
\setlength{\textfloatsep}{0pt}
}


\begin{figure*}[ht]
\setlength{\abovecaptionskip}{0.cm}
\centering
{
\includegraphics[width=0.95\linewidth]{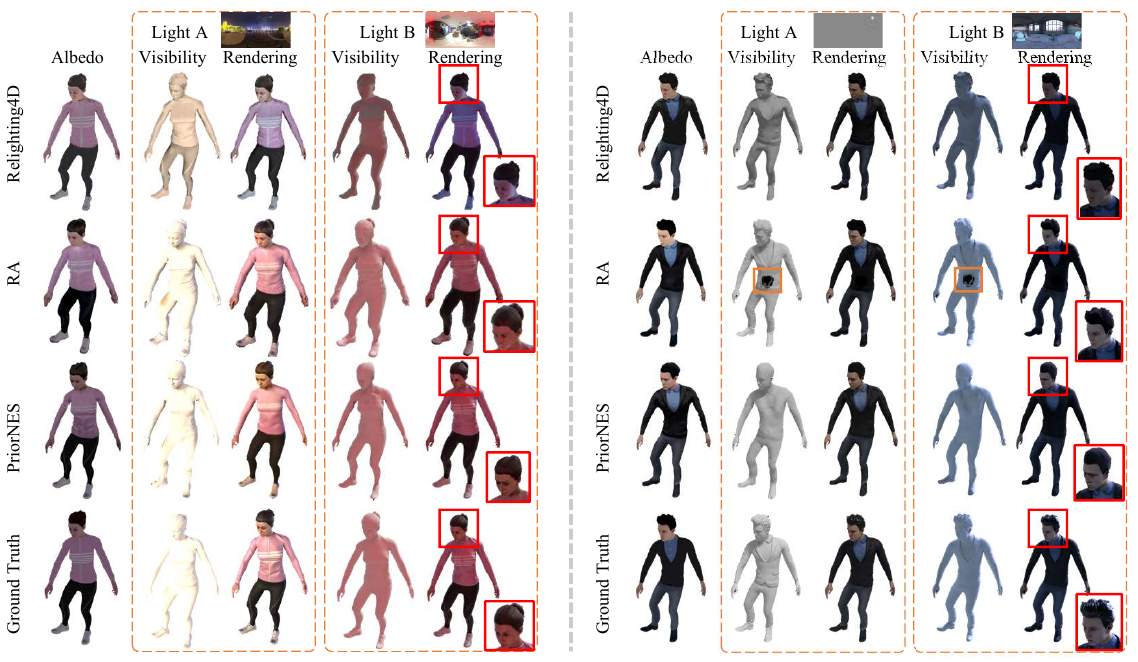}
}\caption{Qualitative Comparisons with RA and Relighting4D on albedo, visibility, and rendering under four novel lights on two different synthetic humans.}
\vspace{-0.6cm}
\label{fig:relighting}
\setlength{\textfloatsep}{0pt}
\end{figure*} 

We assess mesh reconstruction quality using the SyntheticHuman Dataset, comparing NeuralBody (NB)~\cite{peng2021neural}, AniSDF, Relighting4D, and RA.
The average results for all seven human models are presented in Table~\ref{table:mesh}. Additionally, Fig.~\ref{fig:normal_qual} provides a qualitative visualization of the reconstructed normal map for the model \textit{Josh}.
PhyNES outperforms NB, Relighting4D, and AniSDF quantitatively by learning SMPL surface deformations effectively, enhancing dynamic modeling capabilities.
However, PhyNES falls short of RA's performance, which benefits from a more accurate SDF but comes with a training time of over 50 hours compared to PhyNES's 5 hours.
Fig.~\ref{fig:normal_qual} shows that NB generates noisy geometry due to a lack of surface constraints, while AniSDF and RA produce smoother surfaces with signed distance fields as representation.

\subsection{Evaluation on Relighting Applications}
\label{QuantitativeRelighting}

\begin{table}[h!]\footnotesize
    \setlength{\abovecaptionskip}{-0cm}
\setlength{\belowcaptionskip}{-0.4cm}

\begin{center}
    \caption{Comparisons of diffuse albedo, visibility relighting, and speed. Best results are highlighted as \gold{1st} and \silve{2nd}. 
    }
    
    \resizebox{1.0\linewidth}{!}{
    \begin{tabular}{c|c|c|c|c|c|c|c|c}

    \multicolumn{1}{c|}{}   &  \multicolumn{2}{c|}{\textbf{Diffuse Albedo}} & \multicolumn{2}{c|}{\textbf{Visibility}} &  \multicolumn{2}{c|}{\textbf{Relighting}} & \multicolumn{2}{c}{\textbf{Speed}} \\
         & PSNR $\uparrow$ & LPIPS$^*$ $\downarrow$ & PSNR $\uparrow$ & LPIPS$^*$ $\downarrow$ & PSNR $\uparrow$ & LPIPS$^*$ $\downarrow$& Training $\downarrow$ & Rendering $\uparrow$    \\ 
    \hline
    NeRFactor (1 frame) & 22.23&   226 & 11.37 &  387 &   21.04  & 313  & 50+ hours & 0.3- FPS\\
    Relighting4D               & \gold{36.15}               &  29.43&    26.29             & 20.96                  &  33.58          &   24.05             &   \silve{40+  hours}      & 0.3- FPS \\
    RA                         & 35.24& \silve{28.91}& \gold{35.10}  & \gold{16.54}    & \silve{35.30}          & \silve{21.53}          & 50+ hours & \silve{0.5- FPS}   \\
    Ours                       & \silve{35.25}& \gold{28.88}& \silve{34.52}           & \silve{20.08}             & \gold{35.33} & \gold{21.28} &  \gold{5 hours} & \gold{12+ FPS }\\
    \end{tabular}
    \label{table:relighting}
    \setlength{\abovedisplayskip}{-1.cm}
}
\end{center}
\end{table}

We conduct a comparative analysis of two leading relightable avatar methods, RA and Relighting4D, along with NeRFactor~\cite{zhang2021nerfactor}, focusing on static objects trained from a single frame as outlined in~\cite{xu2024relightable}. Our evaluation includes albedo maps, visibility maps, quality of rendered novel views under novel lighting conditions (provided as 6 different light probes in the SyntheticHuman++ dataset), and time for training and rendering. Results, summarized in Table~\ref{table:relighting} (Here LPIPS$^*$ denotes LPIPS $\times\ 10^3$.) and further detailed in supplementary materials, included assessments under four novel lighting conditions for two subjects (Fig.~\ref{fig:relighting}). We assess full images and rendering speeds at a resolution of $512\times512$ while results in~\cite{xu2024relightable} focused on foreground images. 

As shown in Table~\ref{table:relighting}, PhyNES surpasses Relighting4D in visibility and relighting and outperforms RA in albedo and relighting, benefiting from the accurate body shape and facial details provided by the SMPL model. Besides, PhyNES is more efficient in training and rendering, crediting our rasterization strategy. 
Overall, our approach shows notable improvements in performance while maintaining low training costs compared to other methods.
Fig.~\ref{fig:relighting} shows that Relighting4D generates irregular baked-in visibility colors and inaccurate reflection components due to its neural network's poor geometric predictions. 
While PhyNES shows slightly lower visibility than RA due to SMPL topology limitations, it significantly enhances relighting visualizations, especially in capturing fine facial details, indicated by red rectangles of Fig.~\ref{fig:relighting}. In contrast, RA often produces blurred facial features because its material network, designed to reduce shadows, inadvertently removes texture details.
Furthermore, RA often renders distorted body shapes, such as fuller faces, due to implicit sampling, while PhyNES leverages a mesh learned from SMPL to ensure body shape fidelity, leading to better metrics.
Moreover, RA suffers from instability issues, including hollow artifacts, highlighted by red rectangles of Fig.~\ref{fig:relighting}, arising from its unstable SDF estimation from its neural networks. In contrast, PhyNES uses an explicit mesh for stable light visibility calculations, effectively decomposing reflection components to deliver realistic relighting results when maintaining details.

\subsection{Ablation Studies}\label{sec:ablation}
\subsubsection{Hash Encoding MLP}
We conduct an ablation study on the Synthetic Human Dataset to assess the effectiveness of the hash-encoding MLP~\cite{muller2022instant}, which replaces the original MLP in NES. The average training time for 500 steps with the hash-encoded MLP is 3 minutes, compared to 10 minutes for the original NES. This result demonstrates that hash encoding significantly accelerates training in PhyNES.

\subsubsection{Mesh Loss}
We enhance NES by introducing a mesh loss $l_{mesh}$ to improve the mesh quality. 
We quantitatively assess the mesh quality on the Synthetic Human Dataset, with average results presented in Table~\ref{table:mesh_loss}. PhyNES (W/O $l_{mesh}$) denotes PhyNES without $l_{mesh}$ the same as NES. PhyNES (W $l_{mesh}$) denotes PhyNES with $l_{mesh}$ which is used as the base method in this work. Results show NES with $l_{mesh}$ outperforms the original NES, as evidenced by the reductions in both CD and P2S after incorporating $l_{mesh}$. 

\vspace{-0.4cm}
\begin{table}[ht]\footnotesize
\setlength{\abovecaptionskip}{0.cm}

\begin{center}
    \caption{Ablation Study on Mesh Loss. The \textbf{best} results are highlighted in bold.} 
    \scalebox{0.9}{
    \begin{tabular}{c|c|c}
    P2S$\downarrow$ / CD$\downarrow$ &    PhyNES (W/O $l_{mesh}$)         &    PhyNES (W $l_{mesh}$)\\
    \hline
    
    \hline 
    Avg.   & 0.54 / 0.68  & \textbf{0.50} / \textbf{0.62} \\

    \end{tabular}
    
}
    \label{table:mesh_loss}
\setlength{\textfloatsep}{0pt}
    
\end{center}
\end{table}

\vspace{-0.6cm}
\section{Conclusion}
\label{sec: conclusion}

We introduce PhyNES, which represents human avatars using neural maps derived from an implicit signed distance field (SDF). By creating a differentiable connection between the SDF and the body mesh, PhyNES effectively merges the benefits of both implicit and explicit techniques, allowing for efficient geometry inference in a compact transformed space based on a parameterized linear human shape model. The dynamic geometry, texture, and material maps are represented as 2D neural maps, enabling efficient queries for all on-screen pixels via a neural rasterizer. Additionally, PhyNES simplifies material attribute acquisition through interactions with various illumination setups, enabling real-time physically-based rendering. Experiments show that PhyNES achieves comparable relighting qualities to existing SOTA methods while significantly enhancing reconstruction quality, rendering speed, and memory efficiency.

\section*{Acknowledgments}
This research was supported by the Theme-based Research Scheme, Research Grants Council of Hong Kong (T45-205/21-N), and the Tier-2 Start-up Grant, Hong Kong Baptist University (RC-OFSGT2/20-21).


\bibliographystyle{IEEEtran}

\bibliography{myRefs}
\clearpage
\end{document}


\title{Supplementary Material for: 
\\Fast and Physically-based Neural Explicit Surface for Relightable Human Avatars}

\author{
	\textit{Jiacheng Wu$^{1*}$, Ruiqi Zhang$^{1*}$, Jie Chen$^{1\dag}$, Hui Zhang$^{2}$}\\
	$^1$Department of Computer Science, Hong Kong Baptist University, Hong Kong SAR, China \\
    $^2$Department of Computer Science, Beijing Normal-Hong Kong Baptist University, 
    Guangdong, China \\ 
    \tt\small Email: \{csjcwu, csrqzhang, chenjie\}@comp.hkbu.edu.hk, amyzhang@uic.edu.cn \\
    \thanks{$^*$Equal contribution. $^\dag$Corresponding author.}
}

\maketitle
\renewcommand{\thefootnote}{}




\section{Network Architecture}

\begin{table*}[ht!]
\begin{center}
    \caption{Quantitative results of seen pose and novel pose animation on the ZJU-MoCap dataset. We evaluate results using PSNR and LPIPS. The \gold{best} and \silve{second} results for each experiment data have been highlighted accordingly.}
    \label{tab:zjumocal_nvnp}
    \resizebox{0.95\textwidth}{!}{
    \begin{tabular}{c|c|c|c|c|c|c|c|c|c|c|c|c}
        & \multicolumn{6}{c|}{Seen Pose Rendering (PSNR $\uparrow$ / LPIPS $\times 10^3 \downarrow$)} & \multicolumn{6}{c}{Novel Pose Rendering (PSNR $\uparrow$ / LPIPS $\times 10^3 \downarrow$)}\\
    
        ID    & AniSDF &  NDF  & SANeRF & RA & NES & OURS           & AniSDF &  NDF  & SANeRF & RA & NES & OURS \\
        
    \hline
    377 & \gold{27.44} / 21.41  & 26.84 / 23.23  & \silve{27.13} / \gold{19.01} &  26.77 / 20.67 & 27.01 / 19.62 & 27.02 / \silve{19.77} & \gold{25.14} / 23.51 & 24.59 / 26.32 &  24.87 / \gold{22.22} &  25.00 / 23.12 & \silve{25.13} / 22.74 &    25.12 / \silve{22.72} \\
    
    386 & \silve{28.29} / 27.66 & 28.21 / 22.05 & \gold{28.58} / \silve{18.10}   & 28.01 / 29.76 & 28.07/ 22.27 &  28.10 /  \gold{18.09}   & 26.77 / 32.19 & 26.32 / 28.93 &  26.43 / \gold{24.89} &   \silve{26.81} / 31.87 & \gold{27.20} / \silve{27.48} & 27.18 / 27.51   \\
    
    387 & 25.71 / 42.15         & 24.52 / 37.31 & 25.93 / 30.65   & \silve{26.09} / \silve{28.71}  & 26.06 / 26.96 & \gold{26.30} /\gold{26.52} & 23.33 / 43.73 & 22.40 /  40.73 &  22.84 / 36.85 &  \silve{23.35} / \silve{35.71} & \gold{23.60} / 29.75 & 23.59 / \gold{29.33} \\
    
    392 & 28.81 / 32.95  & 28.39/ 29.87 & \gold{29.03}/ \gold{22.56}  &  \silve{28.91} / \silve{31.24} & 28.09 / 23.90 &  28.08 / 23.93   & \silve{25.66} / 39.96  & 25.61 / 39.05 &  25.49 / \silve{33.03} &  25.22 / 41.89 & 26.26 / 31.06 & \gold{26.60} / \gold{29.76}  \\
    
    393 & 26.74 / 33.11         & 26.73 / 30.98 & \gold{27.18} / \gold{25.56}  & 26.75 / 26.77 & 26.78 / 25.63 &  \silve{26.79} / \silve{25.66}  & 23.47 / 44.64 & \silve{24.02} / 42.66 &  \silve{24.02} / \silve{39.10} &  23.97 / 42.65 & 24.50 / 38.60 & \gold{24.51} / \gold{38.57} \\
    \hline
    Avg.& \silve{27.39} / 31.46 & 26.94 / 28.68 & \gold{27.56} / \silve{23.18} & 27.31 / 27.43 & 27.20 / 23.68 &  27.26 / \gold{22.79}  & \silve{24.90} / 36.80 & 24.58 / 35.53& 24.72 / \silve{31.21}  & 24.87 / 35.05 & 25.34 / 29.93 & \gold{25.40} / \gold{29.58}             \\
    \hline
    
    \end{tabular}
    }
    \label{table:novel_quan_all}
\end{center}
\end{table*}

\begin{figure*}[ht!]
\centering
\includegraphics[width=0.98\linewidth]{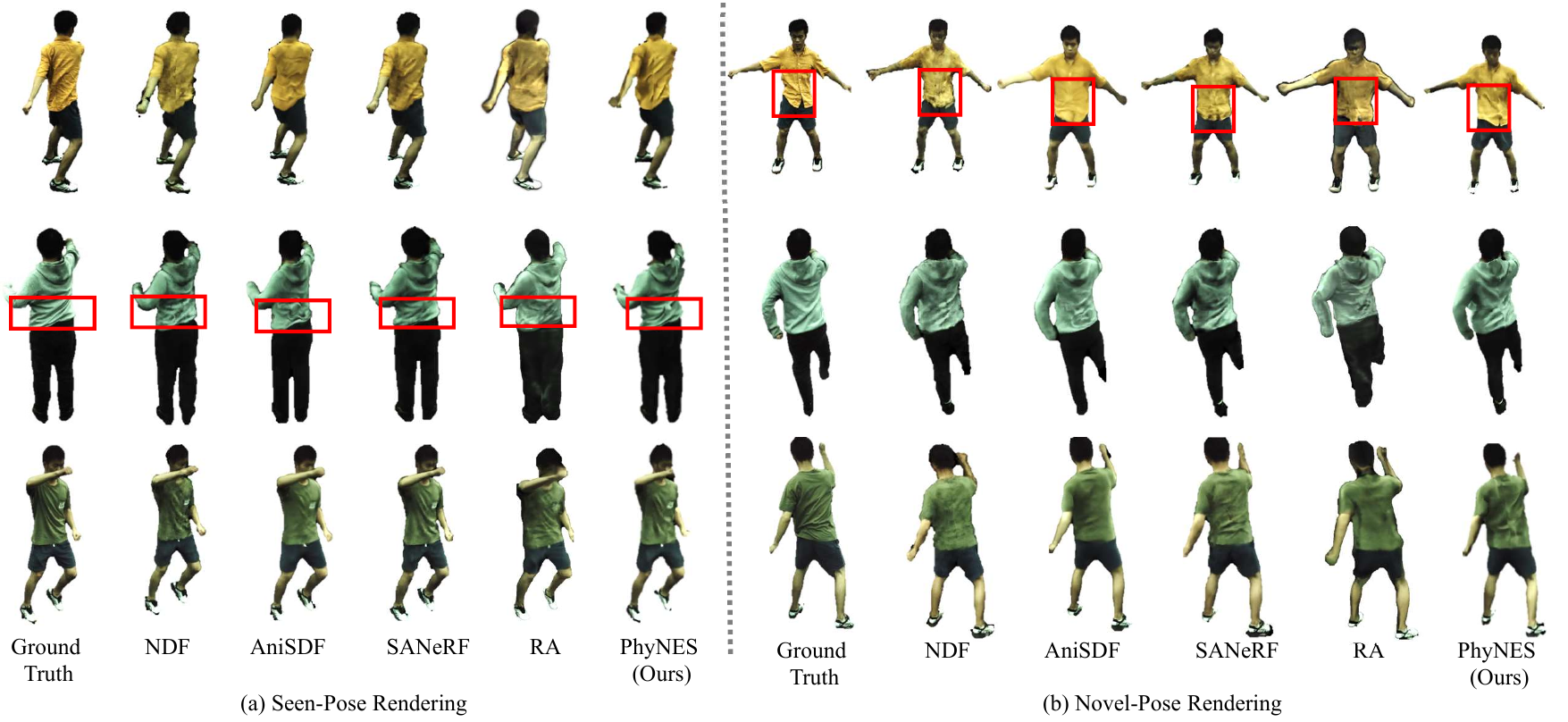}
\caption{Qualitative comparison in (a) seen pose rendering task, and (b) novel pose rendering task on the ZJU-MoCap dataset. As indicated in red circles, our PhyNES present clearer wrinkles and folders than other methods.}
\label{figure:novel_view_all}
\end{figure*}

\begin{table}[ht!]
\begin{center}
    \caption{Quantitative Results on Mesh Reconstruction on Synthetic Human Dataset. The \gold{best} and \silve{second} best results for each experiment data have been lighted accordingly.}
    \vspace{-0.2cm}
    \resizebox{0.9\linewidth}{!}{
    \begin{tabular}{c|c|c|c|c|c}
   P2S$\downarrow$ / CD$\downarrow$ & Neural Body &  AniSDF & Relighting4D & RA & Ours \\
    
    \hline
    
    S1       & 1.29  / 1.21  & 0.62 / 0.65 &  1.67 / 1.58   &\gold{0.51} / \gold{0.53} & \silve{0.58} /\silve{0.57} \\
    S2	     & 1.20  / 1.12  & 0.67 / 0.66 &  1.54 / 1.44   &\gold{0.49} / \gold{0.46} & \silve{0.60} /\silve{0.60} \\
    S3	     & 1.60  / 1.67  & 0.93 / 1.24 &  1.77 / 1.61   &\gold{0.56} / \gold{0.67} & \silve{0.73} /\silve{0.98} \\
    S4       & 0.98  / 1.11  & 0.59 / 0.74 &  1.22 / 1.13   &\gold{0.39} / \gold{0.53} & \silve{0.46} /\silve{0.61} \\
    S5       & 0.98  / 0.99  & 0.42 / 0.54 &  1.25 / 1.21   &\gold{0.35} / \gold{0.45} & \silve{0.39} /\silve{0.50} \\
    S6       & 0.87  / 1.02  & 0.54 / 0.78 &  1.10 / 1.21   &\gold{0.34} / \gold{0.54} & \silve{0.43} /\silve{0.63} \\
    S7       & 0.80  / 0.99  & 0.35 / 0.50 &  1.07 / 1.10   &\gold{0.27} / \gold{0.44} & \silve{0.30} /\silve{0.47} \\
    \hline
    Avg.     & 1.10 / 1.16   & 0.59 / 0.73 &  1.37 / 1.33  &\gold{0.42} / \gold{0.52} & \silve{0.50} / \silve{0.62}\\
   
    \end{tabular}
    }
    \label{table:mesh_all}
\end{center}
\end{table}
\subsection{Quantitative Evaluation on Relighting}

\begin{table}[ht]\scriptsize
\begin{center}
    \caption{Additional Study on Mesh Split:  PhyNES (W Mesh Split) denotes the PhyNES with split mesh as the initialization of training. PhyNES (W/O Mesh Split) denotes the PhyNES without mesh split that is used as the base method in this work. The \textbf{best} results have been bolded.}
    \resizebox{1.0\linewidth}{!}{
    \begin{tabular}{c|c|c|c}

     & Training Time$\downarrow$ & P2S$\downarrow$ & CD$\downarrow$ \\
 \hline
     
 PhyNES (W Mesh Split)  &  50 minutes  & 0.51  &  0.63 \\
    \hline
PhyNES (W/O Mesh Split) &  \textbf{3 minutes} & \textbf{0.50} & \textbf{0.62}  \\

    \end{tabular}
    }
    \label{table:mesh_split}
\end{center}
\end{table}

\begin{table*}[ht!]\footnotesize
\caption{Additional Quantitative Results on All Characters of SyntheticHuman++ Dataset. This table compares our method with baselines on the SyntheticHuman++ dataset~\cite{peng2024animatable}. The \gold{best} and \silve{second} results for each experiment data have been highlighted.}
\begin{center}
    \begin{tabular}{c|c|c|c|c|c|c|c}
    \multicolumn{1}{c|}{} & \multicolumn{1}{c|}{}   &  \multicolumn{2}{c|}{\textbf{Diffuse Albedo}} & \multicolumn{2}{c|}{\textbf{Visibility}} &  \multicolumn{2}{c}{\textbf{Relighting}}
\\
ID & Method & PSNR $\uparrow$ & LPIPS  $\downarrow$ & PSNR $\uparrow$ & LPIPS  $\downarrow$ & PSNR $\uparrow$ & LPIPS $\downarrow$\\ 
\hline
jody & NeRFactor (1 frame)  &   17.74   & 201   & 16.12   & 283    & 21.66  & 224 \\           
     & Relighing4D        & \gold{34.02}   & \silve{34.73}& 24.09         & 29.99             & 30.14          & 31.04               \\
     & RA  & 33.03& 34.91& \gold{33.69}  & \gold{22.22}      & \silve{33.41}  & \silve{27.55}       \\
     &  Ours              & \silve{33.25}& \gold{34.56}& \silve{32.52} & \silve{24.65}        &\gold{33.59}    & \gold{26.59}        \\
\hline 
josh & NeRFactor (1 frame)  &   29.14   & 167   & 14.85   & 291    & 24.68  & 228 \\ 
     & Relighing4D         &  \gold{35.42}        &  48.98           &  23.67         & 32.16        & 31.53          & 39.71           \\
    & RA   & \silve{33.78}      & \silve{45.87}& \gold{32.19}  & \gold{26.36} & \silve{33.29}  & \silve{32.83}   \\
    &  Ours               & 33.69& \gold{44.56}& \silve{31.67} & \silve{29.86}   &\gold{33.32}    & \gold{32.56}   \\
\hline
megan & NeRFactor (1 frame)  &   15.11   & 394   & 7.41  & 538    & 14.06   & 472 \\ 
     & Relighing4D           &  \gold{38.95}  &  \gold{14.06}  &   27.94        & \silve{10.72}     & 33.86          &  \silve{11.79}     \\
     & RA      & \silve{36.40} & \silve{14.09}  & \gold{37.01}   & \gold{8.79}       & \gold{35.97}   & \gold{11.61}     \\
     &  Ours                & 35.95& 15.30& \silve{36.86}	& 12.05             & \silve{35.95}  & 12.71               \\
\hline
leonard & NeRFactor (1 frame)  &   26.93   & 143   & 7.09   & 435    & 23.77  & 326 \\  
    & Relighing4D         &  \gold{36.23}   & \silve{47.39}  &  23.46        & 32.70             & 33.54       & 39.06                    \\
     & RA   & 33.09& \gold{46.57}  & \gold{32.98}  & \gold{26.90}        & \gold{35.69}       & \gold{32.86}    \\
     &  Ours            & \silve{33.27}& 47.40& \silve{31.96} & \silve{30.65}     & \silve{35.65} & \silve{33.57}	     \\
\hline
malcolm & NeRFactor (1 frame)  &   -   & -   & -   & -    & -  & - \\ 
    & Relighing4D             &   \gold{42.19}  & \gold{8.32}&   32.20        & \gold{4.86}      & 38.38         & \gold{7.91}        \\
     & RA       & \silve{39.04}& 9.30& \gold{38.15}  & \silve{5.05}      & \silve{39.75} & 8.36               \\
     &  Ours                & 38.94& \silve{8.56}& \silve{37.56} & 10.98                &\gold{39.77}   & \silve{8.21}    \\
\hline
nathan & NeRFactor (1 frame)  &   -  & -   & -   & -    & -  & - \\ 
    & Relighing4D             & 30.09            &  23.11&  26.35         & 15.28              & \gold{34.03}   & \silve{14.77}            \\
     & RA       & \silve{36.10}& \gold{22.70}   & \gold{36.59}   & \gold{9.90}        & 33.66         & 15.94              \\
     &  Ours             	   & \gold{36.41}& \silve{22.90}& \silve{36.54}  & \silve{12.26}       & \silve{33.67}  & \gold{14.01}            \\
\hline
    \end{tabular}
    \end{center}
    \label{tab:relighting_all}
\end{table*}

The surface geometry network $M_{geo}$ takes two inputs: texel coordinates $(u, v)$ and the pose $\theta$, to output the offset $l$. 
We first encode the input using hash encoding~\cite{muller2022instant} and use 3-layer MLPs, composed of three fully connected layers with SoftPlus activation function applied to each layer. A Residual Fully Connected Layer is used to process the final output.

The dynamic texture network $M_{dtx}$ is designed to process inputs comprising 2D coordinates $(u, v)$ and pose $\theta$ to yield color $c$. We adopt 5-layer MLPs for $M_{dtx}$, composed of five fully connected layers with ReLU activation functions applied to each layer. A Sigmoid activation function is utilized for transforming the final output.


The albedo and roughness networks $M_{alb}$ and $M_{rgh}$ are structured to process inputs consisting of 2D coordinates $(u, v)$ and pose $\theta$, aiming to produce outputs of an albedo item $\alpha_s$  and a roughness item $\gamma_s$. 
We use 5-layer MLPs for $M_{alb}$ and $M_{rgh}$, composed of five fully connected layers with Softplus activation functions applied to each layer. The final output transformation is facilitated by using Softplus and Sigmoid activation functions.

\section{Loss Functions}

\subsubsection{First-stage Training}
Since we divide the PhyNES into two-stage training, we adopt the same training strategy and loss item in NES~\cite{zhang2023explicifying} to optimize the first-stage training.
In the first training stage, a single-stage sampling strategy with 128 samples and an additional 16 points around the first ray intersection is used to maintain accurate surface estimation. This rendering pixel color is obtained by the standard volumetric rendering process:

\begin{align}
   C(r) = \sum_{k=1}^N (\prod_{m=1}^{k-1}e^{-\sigma_m \cdot \delta_m})(1-e^{-\sigma_k \cdot \delta_k}) \cdot c_k,
\end{align}

where $N$ denotes the number of sample points along a ray $\bm{r}$. $c_k$ denotes the predicted color for a sample point. $\delta_k$ denotes the quadrature segment. $\sigma$ is the converted volume density from the offset $\bm{l}$. 

The $M_{geo}$ and $M_{rgh}$ are optimized by MSE loss of pixel colors, denoted as:

\begin{align}
    L_{MSE} = \frac{1}{n} \sum_{i=1}^{n} (c_i - \hat{c}_i)^2,
\end{align}

where $c_i$ represents the true target value for the ith sample,
$\hat{c}_i$ represents the predicted value for the ith sample.

We adopt the mesh loss $l_{mesh}$, including mesh edge loss, mesh normal smoothness loss, and mesh Laplace loss to regularize the mesh's smoothness. We construct the deformed mesh to calculate these losses using Pytorch3d. The mesh edge loss is used to penalize deviations in the edge lengths of a mesh. The mesh normal smoothness loss promotes smoothness in the normal vectors across adjacent faces. The mesh Laplace loss encourages smoothness in vertex positions based on the Laplacian operator. We describe this mesh loss as:

\begin{align}
    L_{mesh} &= w_{edge}\cdot L_{edge}\notag
    \\ &+ w_{normal\ smoothness}\cdot L_{normal\ smoothness}\notag
    \\ &+ w_{Laplace}\cdot L_{Laplace},
\end{align}

where $w_{edge}$, $w_{normal\ smoothness}$, and $w_{Laplace}$ denote the weight of applying different loss items respectively. In our experiment, we set $w_{edge}$ = 1, $w_{normal\ smoothness}$ = 0.01, and $w_{Laplace}$ = 0.1.

\subsubsection{Second-stage Training}
In the second training stage, we fix $M_{geo}$ and $M_{rgh}$ when training our material network $M_{alb}((u, v), \theta) = \alpha_s$ and $M_{rgh}((u, v), \theta) = \gamma_s$. We optimize PhyNES by rendering the image with given camera poses and comparing the pixel values $L_o$ obtained from the rendering equation against the ground truth ones $L_{gt}$, The main loss function is given by:

{\footnotesize
\begin{align}
    L_{material} = \sum_{r\in R} \|L_{gt}(r) - L_o(r)\|_2,
\end{align}
}

where $\bm{r} = o + \bm{t}d \in R$ denotes the camera rays in the rendering process.

Following~\cite{chen2022relighting4d}, we add smoothness and sparsity regularization to our material network to alleviate ambiguities. The regularization terms can be described as:

{\footnotesize
\begin{gather}
L_{\alpha\!Entropy}\!=\!GaussianEntropy(M_{alb}((u, v), \theta)),
\\L_{\alpha_s}\!=\!\sum_{(u, v)
\atop\in(U, V)}\!\|M_{alb}((u, v), \theta)\!- M_{alb}((u+\Delta u, v+\Delta v), \theta)\|_2,
\\L_{\gamma_s}\!=\!\sum_{(u, v)
\atop\in(U, V)}\!\|M_{rgh}((u, v), \theta)\!-\!M_{rgh}((u+\Delta u, v+\Delta v), \theta)\|_2,
\end{gather}
}

where $GaussianEntropy(M_{alb}((u, v), \theta))$ means the entropy of the Gaussian distribution of the albedo map $\alpha_s$ evaluated on the $(u, v)$ space. $(\Delta u$, $\Delta v)$ are random noise sampled from a normal distribution $(\Delta u$, $\Delta v)\!\sim\!N(\mu, \sigma^2)$. We follow~\cite{xu2024relightable} to set $\mu$ as 0 and $\sigma^2$ as 0.02 in our experiments.
 
\section{Additional Experiment Results}
\subsection{Rendering Quality Evaluation}
More quantitative comparisons are shown in Table\ref{table:novel_quan_all} and qualitative comparisons in Fig.~\ref{figure:novel_view_all}. We follow the data preprocessing method and the training-test split of~\cite{peng2024animatable} for the ZJU-MoCap dataset.

\subsection{Mesh Reconstruction Quality}
We show more quantitative results on mesh reconstruction in Table~\ref{table:mesh_all}. We follow the data preprocessing method and training setting of ~\cite{peng2024animatable} for the SyntheticHuman dataset.

\subsection{Relighting Quality Evaluation}
We report more quantitative results on relighting performed in the SyntheticHuman++ dataset in Table~\ref{tab:relighting_all}. We follow the data preprocessing method and configuration of~\cite{xu2024relightable} for the SyntheticHuman++ dataset. To mitigate scale ambiguity in inverse rendering, we aligned predicted images with ground truth using channel-wise scale factors based on methodologies from~\cite{xu2024relightable}. 

\section{Additional Experiment: Mesh Split}
Since the vertices and faces of the SMPL are much smaller than the artist-created meshes of the Synthetic Human Dataset, it is reasonable to assume that more vertices and faces of the original body model can provide more possibilities in fitting with human models. Therefore, we split the SMPL in the center of each triangle on the mesh to increase the vertices from 6890 to 27554 and faces from 13776 to 55104. We choose seven humans in the Synthetic Human Dataset to evaluate and show the average values. We used CD, P2S in \textit{cm}, and training time to quantitatively evaluate the mesh after adding these losses in the Synthetic Human Dataset. To test the training time, we train each model for 40 epochs and calculate the time cost for each epoch in minutes on a machine with one 32G NVIDIA V100. The comparison is shown in Table~\ref{table:mesh_split}. However, results show that training starting from the split SMPL will not improve CD and P2S but will greatly increase the training burden. So, we still use the original SMPL for training.

\section{Limitations and Future Directions}\label{sec: limitations}

There are several limitations to our PhyNES. Firstly, the details of the hair and fingers are difficult to predict, and the mesh of our method is limited to the topology of the SMPL. We plan to adopt an efficient remeshing strategy in certain regions to enhance the details and break through the constraints from SMPL's topology. Secondly, the texture map contains the wrinkle information which should be represented in the geometry, which results in the albedo and degree of normal of our method being worse than RA. We are researching a method to enhance the communication between texture and geometry to eliminate this problem. Thirdly, the properties of the dynamic material are anti-intuitive. Although our final material maps are fixed even with different poses, there is some space to improve the training of material properties. We are looking for an approach to fix the texture map obtained in the first stage as the based albedo map in the second stage to integrate these two stages.

\bibliographystyle{IEEEtran}
\bibliography{myRefs}
\clearpage